\documentclass[pdflatex,sn-nature]{sn-jnl}

\usepackage{graphicx}
\usepackage{multirow}
\usepackage{amsmath,amssymb,amsfonts}
\usepackage{amsthm}
\usepackage{mathrsfs}
\usepackage[title]{appendix}
\usepackage{xcolor}
\usepackage{textcomp}
\usepackage{manyfoot}
\usepackage{booktabs}
\usepackage{tabularx}
\usepackage{algorithm}
\usepackage{algorithmicx}
\usepackage{algpseudocode}
\usepackage{listings}
\usepackage{subcaption}
\usepackage{graphicx}
\usepackage{booktabs}
\usepackage{rotating}

\theoremstyle{thmstyleone}

\theoremstyle{thmstyletwo}

\theoremstyle{thmstylethree}

\begin{document}

\title[Article Title]{DT-ICU: Towards Explainable Digital Twins for ICU Patient Monitoring via Multi-Modal and Multi-Task Iterative Inference}

\author{\fnm{Wen} \sur{Guo}}\email{guowen0903@hotmail.com}

\affil{\orgname{ETH Zurich}, \orgaddress{\country{Switzerland}}}

\abstract{
We introduce DT-ICU, a multimodal digital twin framework for continuous risk estimation in intensive care. DT-ICU integrates variable-length clinical time series with static patient information in a unified multitask architecture, enabling predictions to be updated as new observations accumulate over the ICU stay. We evaluate DT-ICU on the large, publicly available MIMIC-IV dataset, where it consistently outperforms established baseline models under different evaluation settings.

Our test-length analysis shows that meaningful discrimination is achieved shortly after admission, while longer observation windows further improve the ranking of high-risk patients in highly imbalanced cohorts. To examine how the model leverages heterogeneous data sources, we perform systematic modality ablations, revealing that the model learnt a reasonable structured reliance on interventions, physiological response observations, and contextual information. These analyses provide interpretable insights into how multimodal signals are combined and how trade-offs between sensitivity and precision emerge.

Together, these results demonstrate that DT-ICU delivers accurate, temporally robust, and interpretable predictions, supporting its potential as a practical digital twin framework for continuous patient monitoring in critical care.
}

\maketitle
\section{Introduction}\label{sec_intro}

The concept of digital twins originates in the domain of engineering, where it was introduced to create continuously updated virtual counterparts of physical systems for real-time monitoring, simulation, and predictive maintenance\cite{datta2016emergence,aivaliotis2019use}, a paradigm that is now being adapted to model complex biological and clinical systems in healthcare domains\cite{elgammal2025digital, dervisoglu2023cloudEdgeDT,
shamanna2020hba1cDT,
abirami2023dths,
abilkaiyrkyzy2024dialogueDT,
sarp2023woundDT,
avanzato2024lungdt,
kolekar2023precisionDT,
kumi2024stressDT,
wang2024twingpt,
makarov2024llmdt,
moore2024syntwin, dong2021machine}.
As artificial intelligence becomes more deeply embedded in clinical workflows, there is growing interest in moving beyond static prediction models toward systems that can continuously track patient state, anticipate clinical deterioration, and support timely decision making. In this context, digital twins offer a promising direction: they are not only passive predictors, but evolving models of individual patients that can be updated as new data arrive. In high-acuity settings such as the intensive care unit (ICU), where conditions change rapidly and treatment decisions are time-critical, this ability to maintain a continuously updated view of the patient is especially important.

For a digital twin to be clinically meaningful in healthcare, several key capabilities are required. First, it must integrate heterogeneous clinical data, including structured time-series inputs (e.g., vital signs and laboratory measurements) together with static patient information (e.g., demographics and diagnoses). Second, it must support continuous updating, so that predictions and internal representations evolve as new observations become available. Third, it should enable systematic perturbation or “what-if” analyses, allowing clinicians and researchers to examine how changes in available information or treatment-related signals may influence predicted outcomes.

These requirements are particularly salient in the ICU, where patient trajectories can shift abruptly and clinicians must make high-stakes decisions under time pressure. Despite the growing use of machine learning models in critical care~\cite{johnson2016machine,alderden2018predicting,meyer2018machine,chen2023machine,sheikhalishahi2020benchmarking}, most existing approaches fall short of this ideal. They are commonly trained and evaluated using fixed-length observational windows, focus on single predictive tasks, and operate in a static inference mode. As a result, such models function primarily as snapshot-based risk calculators, rather than continuously evolving representations of patient state.

A wide range of deep learning methods, including MLP~\cite{he2016deep}, recurrent neural networks (RNNs) \cite{elman1990finding}, long short-term memory (LSTM) models \cite{hochreiter1997lstm}, and more recent transformer-based architectures \cite{vaswani2017attention}, have been applied to ICU prediction tasks such as mortality risk and length-of-stay estimation~\cite{lovon2024revisiting, gupta2022extensive, mu2025predicting,li2025machine,olang2025artificial,pang2022establishment,lin2021empirical}. While these approaches have demonstrated strong performance under controlled settings, they are typically developed for narrow predictive targets and evaluated under rigid data assumptions. Consequently, they are not designed to support continuous updating, multimodal perturbation analysis, or evolving patient representations over time.

In this work, we propose \textbf{DT-ICU}, a multimodal digital twin framework designed to address these limitations and support continuous outcome prediction in critical care. DT-ICU ingests variable-length patient histories composed of heterogeneous clinical data streams and integrates them through a unified multitask learning architecture that supports both classification (e.g., mortality risk) and regression (e.g., physiological trajectories and future predictions). Crucially, the framework enables iterative inference, allowing systematic examination of how different modalities and clinical signals influence model predictions over time. In addition, DT-ICU incorporates attention-based analysis and structured modality ablations, providing transparent insights into its multimodal reasoning process.

We evaluate DT-ICU through a comprehensive set of experiments on the large, publicly available MIMIC-IV dataset~\cite{johnson2023mimic}. DT-ICU consistently outperforms state-of-the-art mortality prediction baselines and it also achieves superior performance relative to standard backbone architectures, including MLP, RNN, and LSTM variants, within our variable-length and multitask framework. By analyzing performance as patient history accumulates, we show that DT-ICU attains stable and reliable discrimination early in the ICU stay while continuing to refine the ranking of high-risk patients over longer horizons.
Beyond aggregate accuracy, we conduct systematic modality ablation studies that expose how different clinical signals are organized within the model, revealing a learned hierarchy in which interventions and physiological responses carry the strongest predictive value. Together, these analyses demonstrate that DT-ICU does not rely on predefined rules about which inputs should matter, but instead learns clinically meaningful multimodal structure directly from data, enabling robust and adaptive digital twin representations that remain effective under heterogeneous, incomplete, and evolving ICU data streams.

\section{Results}\label{sec_results}

\subsection{Robustness to input length and task configuration}
As explained in Section~\ref{sec_intro}, a defining requirement of our ICU digital twin is the ability to work under heterogeneous data availability, and across multiple clinical objectives including detection and regression tasks. In practice, this means that the model must support arbitrary observation lengths after admission and accommodate additional regression tasks without compromising its primary clinical classification performance. We therefore evaluate DT-ICU under different input length and task configurations to assess its robustness as a continuously operating system. Specifically, DT-ICU was tested under three representative settings: (1) a conventional 72-hour fixed-length input without a regression head, (2) variable-length input without a regression head, and (3) variable-length input with multitask prediction. As summarized in Table~\ref{tab:input_settings}, DT-ICU maintains strong and stable performance across all configurations. Firstly, transitioning from a fixed 72-hour window to variable-length inputs does not lead to any degradation in discrimination or classification accuracy, demonstrating that the model can operate effectively at arbitrary time points during the ICU stay. Besides, adding the regression head for multitask learning also does not harm performance on the primary classification task: AUROC and AUPRC remain unchanged between the single-task and multitask settings, while accuracy, precision, and recall show only minor variations. This indicates that DT-ICU can jointly support multiple clinical objectives without inducing negative transfer, confirming its suitability as a unified, continuously operating digital twin.
\begin{table}[ht]
\centering
\caption{Performance under different input length and task configurations, demonstrating robustness to variable-length inputs and multitask prediction.}
\begin{tabular}{lccccc}
\toprule
\textbf{Setting} & \textbf{AUROC} & \textbf{AUPRC} & \textbf{Accuracy} & \textbf{Precision} & \textbf{Recall} \\
\midrule
72h (no regression head) & 0.88 & 0.85 & 0.81 & 0.80 & 0.79 \\
Any length (no regression head) & 0.98 & 0.82 & 0.96 & 0.65 & 0.84 \\
Any length (with regression head) & 0.98 & 0.82 & 0.95 & 0.62 & 0.87 \\
\bottomrule
\end{tabular}
\label{tab:input_settings}
\end{table}

\subsection{Comparison with state-of-the-art and baseline models}
Having established that DT-ICU can operate robustly under flexible input lengths and multitask settings, we next evaluate whether this design translates into superior predictive performance. We conduct two complementary sets of comparisons: (i) against state-of-the-art ICU mortality prediction models reported in the literature under matched clinical settings, and (ii) against standard deep learning backbones under identical data and task configurations, in order to isolate the benefit of the DT-ICU architecture.

\paragraph{Comparison with literature baselines.}
To ensure a fair and clinically meaningful comparison, we evaluate DT-ICU on MIMIC-IV 2.0 using exactly the same data extraction, preprocessing, and evaluation protocols as prior work. In particular, we follow the 72-hour fixed-length input setting and the reported class prevalence (pos.=0.12) used by existing ICU mortality prediction studies. As shown in Table~\ref{tab:mimiciv_sideways} (right block), DT-ICU achieves the highest AUROC and AUPRC among all compared methods, consistently outperforming classical machine learning models, recurrent architectures, and recent literature baselines.

While most prior studies report only AUROC and AUPRC under imbalanced prevalence, one representative work provides an open-source implementation that evaluates models under balanced sampling (pos.=0.5) using five-fold cross-validation. We therefore reproduce this setting exactly and report comprehensive metrics in the left block of Table~\ref{tab:mimiciv_sideways}. Under this more stringent evaluation protocol, DT-ICU again achieves substantially higher AUROC, AUPRC, accuracy, precision, and recall than the best-performing literature baseline, demonstrating robust superiority across both imbalanced and balanced clinical settings.

\paragraph{Comparison with standard backbone models.}
To further disentangle architectural contributions from data and task effects, we additionally compare DT-ICU against commonly used deep learning backbones, including MLP-, RNN-, and LSTM-based models, under the same multitask learning setting on MIMIC-IV 3.1. As summarized in Table~\ref{tab:baselines_iv}, DT-ICU outperforms all baseline architectures across every classification metric while also achieving the lowest prediction error on the regression task.

\begin{table*}[t]
\centering
\footnotesize
\caption{
Comparison with state-of-the-art models on MIMIC-IV 2.0 under two evaluation settings.
All baseline results are taken from Gupta et al.~\cite{gupta2022extensive} unless otherwise specified.
Results of imbalanced setting (pos.=0.12, right block) are taken from published results in prior works~\cite{gupta2022extensive, lovon2024revisiting}.
Balanced setting (pos.=0.5, left block) strictly reproduces the open-source implementation and evaluation protocol of Gupta et al.~\cite{gupta2022extensive}.
}
\label{tab:mimiciv_sideways}

\setlength{\tabcolsep}{3pt}
\renewcommand{\arraystretch}{1.1}

\begin{tabularx}{\linewidth}{@{}>{\raggedright\arraybackslash}X*{7}{c}@{}}
\toprule
{} & \multicolumn{5}{c}{\textbf{Balanced (pos.=0.5)}} & \multicolumn{2}{c}{\textbf{Imbalanced (pos.=0.12)}} \\
\cmidrule(lr){2-6} \cmidrule(lr){7-8}
\textbf{Model} &
\textbf{AUROC} &
\textbf{AUPRC} &
\textbf{Acc.} &
\textbf{Prec.} &
\textbf{Rec.} &
\textbf{AUROC} &
\textbf{AUPRC} \\
\midrule

Logistic Regression & \multicolumn{5}{c}{—} & 0.67 & 0.24 \\
Random Forest       & \multicolumn{5}{c}{—} & 0.79 & 0.39 \\
Gradient Boosting   & \multicolumn{5}{c}{—} & 0.85 & 0.48 \\
XGBoost             & \multicolumn{5}{c}{—} & 0.84 & 0.47 \\
LSTM (TS)           & \multicolumn{5}{c}{—} & 0.86 & 0.49 \\
TCN (TS)            & \multicolumn{5}{c}{—} & 0.84 & 0.46 \\
TCN (Hybrid)        & \multicolumn{5}{c}{—} & 0.85 & 0.47 \\

LSTM (Hybrid)~\cite{gupta2022extensive} &
$0.88\pm0.02$ &
$0.87\pm0.02$ &
$0.80\pm0.02$ &
$0.83\pm0.04$ &
$0.75\pm0.05$ &
0.86 &
0.49 \\

Meditron~\cite{lovon2024revisiting} & \multicolumn{5}{c}{—} & 0.89 & 0.46 \\

\midrule
\textbf{DT-ICU (ours)} &
\textbf{$0.97\pm0.01$} &
\textbf{$0.93\pm0.05$} &
\textbf{$0.96\pm0.01$} &
\textbf{$0.92\pm0.02$} &
\textbf{$0.99\pm0.01$} &
\textbf{0.89} &
\textbf{0.54} \\

\bottomrule
\end{tabularx}
\end{table*}

\begin{table}[ht]
\centering
\caption{
Comparison with different deep learning backbones(MLP, RNN, and LSTM) under the same multitask learning setting on MIMIC-IV 3.1.
}
\begin{tabular}{lcccccc}
\toprule
\textbf{Model} &  \textbf{AUROC} & \textbf{AUPRC} & \textbf{Accurancy} & \textbf{Precision} & \textbf{Recall} & \textbf{MSE} \\
\midrule
MLP-based & 0.88 & 0.51 & 0.93 & 0.51 & 0.63 & 0.05 \\
RNN-based & 0.75 & 0.28 & 0.89 & 0.24 & 0.31 & 0.04 \\
LSTM-based & 0.75 & 0.24 & 0.84 & 0.18 & 0.38 & 0.05 \\
\textbf{DT-ICU (ours)} & \textbf{0.98} & \textbf{0.82} & \textbf{0.95} & \textbf{0.62} & \textbf{0.87} & \textbf{0.02} \\
\bottomrule
\end{tabular}
\label{tab:baselines_iv}
\end{table}

\subsection{Effect of test length}
\label{subsec:ablate_test_length}
A key advantage of DT-ICU is its ability to update predictions continuously as new patient data become available after ICU admission. This raises a critical clinical question: how much temporal context is required before the model produces reliable and clinically useful predictions? To answer this, we evaluate DT-ICU across a wide range of test lengths, from early admission to extended ICU stays, in order to characterize when and how predictive performance stabilizes.

\begin{figure}[t]
  \centering
  \includegraphics[width=\linewidth]{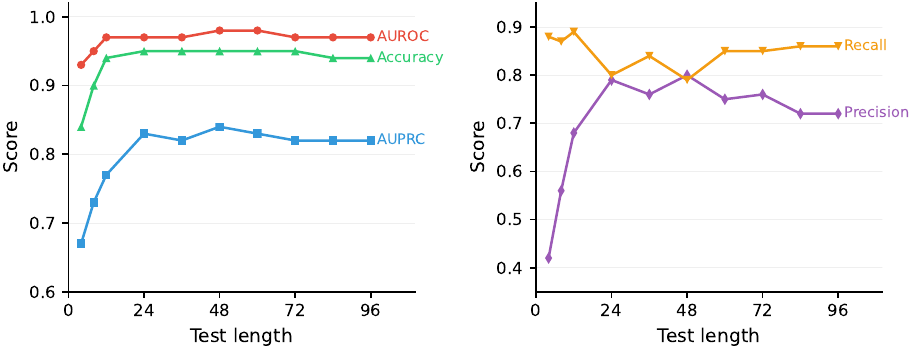}
  \caption{
  Effect of test length on predictive performance.
  The left panel reports AUROC, Accuracy, and AUPRC under different test lengths (hours), while the right panel reports Precision and Recall, highlighting the precision--recall trade-off as the observation window increases.
  }
  \label{fig:test_length}
\end{figure}

Figure~\ref{fig:test_length} summarizes how predictive performance varies as a function of the test length, i.e., the amount of patient history available at inference time.
With respect to discriminative ability, both AUROC and Accuracy exhibit rapid saturation. AUROC increases sharply from 4 to 12 hours and reaches approximately 0.97 by 12–24 hours, after which further gains become negligible. Accuracy follows a similar pattern, stabilizing around 0.95 beyond 24 hours. This indicates that most of the information required to distinguish high-risk from low-risk patients is already present in the early phase of the ICU stay, and that extending the test window beyond one day yields diminishing returns in terms of overall discrimination.
In contrast, AUPRC—which more directly reflects the quality of positive-case ranking in this highly imbalanced clinical setting—continues to improve up to the 24–48 hour range (from 0.67 at 4 hours to approximately 0.84 at 48 hours) and then plateaus. This suggests that, while coarse risk stratification can be achieved early, a longer observation window is beneficial for reliably identifying true positive cases, with 24–48 hours emerging as a practical “sweet spot” for optimizing precision-sensitive clinical detection.
The right panel further reveals a systematic precision–recall trade-off as the test length increases. Precision peaks in the intermediate regime (approximately 24–48 hours, reaching around 0.80), whereas Recall is higher for longer windows (60–96 hours). This reflects a clinically intuitive shift in operating behavior: shorter windows lead to more aggressive predictions with higher precision but lower sensitivity, while longer windows produce more conservative predictions that capture a larger fraction of true cases at the cost of reduced precision. Together, these results highlight that 24–48 hours provides a balanced operating point, combining near-maximal AUPRC with saturated AUROC and Accuracy, while avoiding the diminishing returns and changing clinical trade-offs associated with much longer observation windows.

For completeness and reproducibility, the full numerical results for all test-length settings, including per-metric values and confidence intervals, are reported in Appendix~\ref{sec_appendix_len}.

\subsection{Dealing with data imbalance}
Prediction of ICU outcomes is characterized by extreme class imbalance and highly heterogeneous sequence lengths, both of which can severely distort model optimization and lead to unreliable solutions. While Section~\ref{sec_methods} describes how DT-ICU is designed and trained, we here analyze how its training behavior is shaped by explicit control of these two sources of imbalance.
Specifically, we consider two complementary mechanisms: \emph{sequence-length balancing}, which equalizes the distribution of temporal coverage across patients by balancing input sequence lengths, and \emph{label balancing}, which rebalances positive and negative outcome samples during training. We evaluate four configurations—no balancing, sequence-length balancing only, label balancing only, and their combination—to quantify their individual and joint effects, as summarized in Table~\ref{tab:balancing}. Without any balancing, the model collapses to predicting only the majority class, yielding zero precision and recall despite deceptively high accuracy. Applying either sequence-length balancing or label balancing alone substantially improves performance, but each introduces a characteristic trade-off: sequence-length balancing increases precision at the expense of recall, whereas label balancing increases recall while reducing precision. Combining both strategies yields the best overall performance, achieving simultaneously high AUROC and AUPRC together with a balanced precision–recall profile, while maintaining low prediction error on the regression task. These results demonstrate that DT-ICU’s stability and accuracy under severe class imbalance are enabled by its joint control of temporal and outcome imbalance problem.
\begin{table}[t]
\centering
\small
\caption{Effect of training-time balancing under severe class and sequence-length imbalance in MIMIC-IV.}
\label{tab:balancing}
\setlength{\tabcolsep}{3.5pt}
\begin{tabular}{lcccccc}
\toprule
\textbf{Strategy} & \textbf{AUROC} & \textbf{AUPRC} & \textbf{Acc.} & \textbf{Prec.} & \textbf{Recall} & \textbf{MSE} \\
\midrule
No balancing    & 0.75 & 0.19 & 0.93 & 0.00 & 0.00 & 0.02 \\
Sequence-length balancing only  & 0.96 & 0.73 & 0.96 & 0.78 & 0.55 & 0.02 \\
Label balancing only  & 0.98 & 0.80 & 0.94 & 0.53 & 0.91 & 0.02 \\
Sequence-length + label balancing  & \textbf{0.98} & \textbf{0.82} & \textbf{0.95} & \textbf{0.62} & \textbf{0.87} & \textbf{0.02} \\
\bottomrule
\end{tabular}
\end{table}

\subsection{Multimodal attribution and explainability}
\label{subsec:lomo_ltmo}

To understand how DT-ICU integrates heterogeneous clinical information, we analyze modality attributions and interactions using leave-one-modality-out (LOMO) and leave-two-modality-out (LTMO) ablations, which are commonly adopted in multimodal robustness and cross-modal sensitivity analyses~\cite{mohapatra2024wearable, sharma2025cldm, zhang2025drug}. In both cases, we report performance changes relative to the full-modality baseline, defined as $\Delta = m_{\text{ablated}} - m_{\text{baseline}}$, such that negative values indicate performance drops when the corresponding information is zeroed during inference. Full numerical results are provided in Appendix~\ref{sec_appendix_lomo_ltmo}.

We consider modalities including sequential ICU event data: medication/infusion inputs (\texttt{Meds}), procedures/interventions (\texttt{Proc}), physiological outputs such as urine and drainage (\texttt{Out}), bedside charted measurements (\texttt{Chart}), time-indexed events (\texttt{Date}), and ingredient/component information associated with inputs (\texttt{Ing}), as well as non-sequential patient context from diagnoses (\texttt{Diag}), and demographic data (\texttt{Demo}) which includs gender, age, insurance type, race, icu type, and admission urgency level.

\begin{figure}[t]
  \centering
  \includegraphics[width=0.95\linewidth]{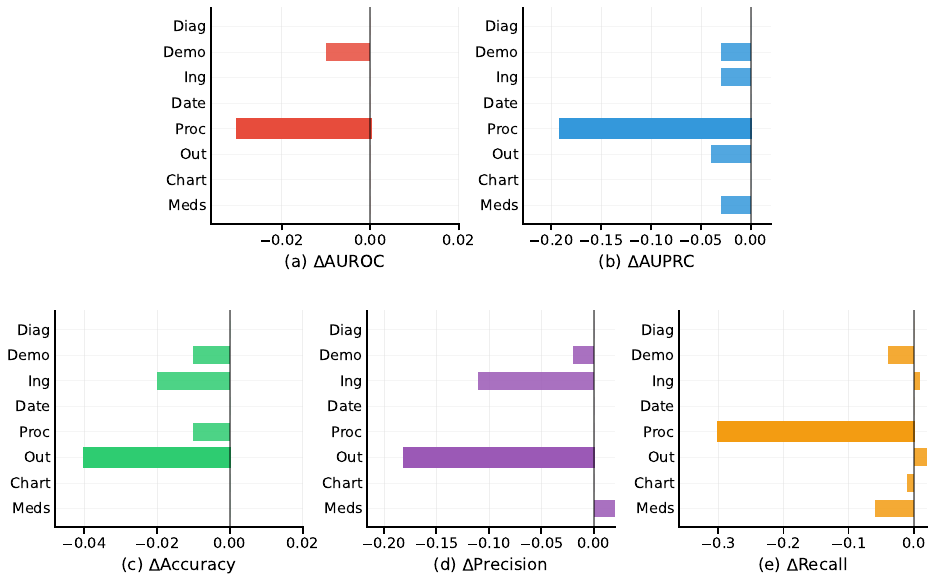}
  \caption{
  Leave-One-Modality-Out (LOMO) ablation analysis showing performance degradation when individual modalities are removed during inference. Each bar represents the change in performance ($\Delta$) relative to the baseline model using all modalities, where negative values indicate performance drops.
  }
  \label{fig:lomo}
\end{figure}

\paragraph{Single-modality attribution (LOMO).}
Figure~\ref{fig:lomo} shows the effect of removing each modality individually, where we could observe strong non-uniform importance. \texttt{Proc} (Procedures) shows the largest overall impact, with AUPRC dropping by 0.19 and Recall by 0.30, indicating that interventional records provide a dominant signal for mortality risk estimation. \texttt{Out} (Outputs) is particularly important for Precision ($\Delta=-0.18$), suggesting that physiological response variables help reduce false positive predictions. \texttt{Chart}, \texttt{Date}, \texttt{Diag}, and \texttt{Demo} exhibit minimal standalone effects, implying that these contextual sources are primarily exploited through interactions with other modalities rather than acting as primary drivers in isolation.

\begin{figure}[t]
  \centering
  \includegraphics[width=\linewidth]{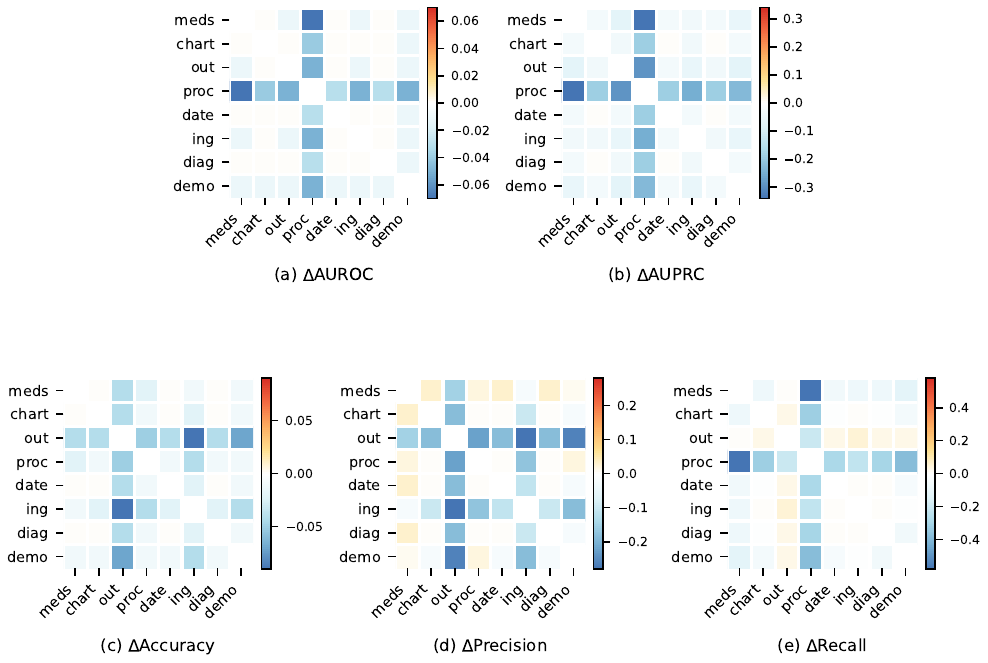}
  \caption{
  Leave-Two-Modalities-Out (LTMO) ablation heatmaps showing performance changes when pairs of modalities are simultaneously removed during inference. Each cell $(i,j)$ represents the change $\Delta$ when both modality $i$ and modality $j$ are zeroed out. Blue intensity indicates severity of performance degradation (darker = worse), white indicates little or no change, and red indicates improvement (rare). Diagonal cells are masked. Panels (a--e) report $\Delta$AUROC, $\Delta$AUPRC, $\Delta$Accuracy, $\Delta$Precision, and $\Delta$Recall changes.
  }
  \label{fig:ltmo}
\end{figure}

\paragraph{Pairwise modality interactions (LTMO).}
While LOMO identifies which modalities matter individually, it does not reveal how they interact. Figure~\ref{fig:ltmo} therefore reports LTMO heatmaps that expose cross-modal dependencies when pairs of modalities are removed simultaneously. The most severe failure modes involve \texttt{Proc} and \texttt{Out}. In particular, removing \texttt{Proc} and \texttt{Meds} together produces the largest degradation ($\Delta$AUPRC $=-0.34$, $\Delta$Recall $=-0.58$), indicating that interventional records and pharmacological inputs jointly encode a substantial fraction of the mortality signal. The combination \texttt{Proc}+\texttt{Out} forms the second most critical failure mode ($\Delta$AUPRC $=-0.29$), appearing as a pronounced ``blue corridor'' across multiple panels, consistent with a tight coupling between delivered interventions and physiological response trajectories.
Across metrics, AUROC remains relatively stable except for \texttt{Proc}-centered pairs, whereas AUPRC and Recall are more sensitive, suggesting that multimodal failures primarily impair positive-case ranking and sensitivity rather than coarse class separation. Taken together, these results support a modality hierarchy in which procedures and outputs form a functional core, medications and ingredients provide complementary treatment-related signals, and date, chart, diagnosis, and demographic features act as contextual inputs.

\begin{figure}[t]
  \centering
  \includegraphics[width=\linewidth]{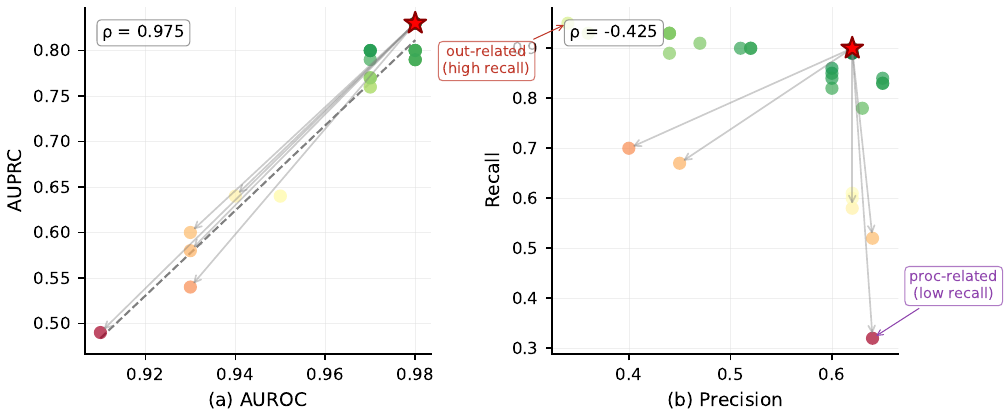}
  \caption{
  Pairwise metric relationships. Each point represents one two-modality zero-out combination (28 pairs across 8 modalities), with the baseline (all modalities) marked for reference. (a)) AUROC and AUPRC show a strong positive correlation (Pearson $\rho=0.975$, $p<0.0001$). (b) Precision and recall have a systematic trade-off pattern (Pearson $\rho=-0.425$, $p=0.024$), with distinct clusters corresponding to different modality-dependent failure modes.
  }
  \label{fig:ltmo_scatter}
\end{figure}

\paragraph{Complementary metric relationships.}
The heatmap shows \emph{which} modality pairs degrade performance for each metric, we further discuss \emph{how} evaluation metrics co-vary across LTMO configurations using a scatter analysis in Figure~\ref{fig:ltmo_scatter}, to better understand which metrics
move together and which trade off against each other when modalities are removed. In the figure, each point corresponds to one of the two-modality zero-out combinations over the 8 modalities (\texttt{Meds}, \texttt{Chart}, \texttt{Out}, \texttt{Proc}, \texttt{Date}, \texttt{Ing}, \texttt{Diag}, \texttt{Demo}), and the baseline works as a reference.
Figure~\ref{fig:ltmo_scatter}(a) shows that AUROC and AUPRC are strongly correlated across all modality pairs ($\rho = 0.975$). When a pair removal hurts AUPRC, it tends to hurt AUROC by roughly the same amount. Figure~\ref{fig:ltmo_scatter}(b) reveals different behavior for Precision and Recall. Pairs involving \texttt{Out} tend to maintain high recall but lose precision, while pairs involving \texttt{Proc} show the opposite—precision holds up better but recall drops sharply. This pattern suggests \texttt{Out} helps the model avoid false alarms, while \texttt{Proc} helps it catch positive
cases. Removing both together degrades performance across the board, consistent with their central importance in Figures~\ref{fig:lomo}--\ref{fig:ltmo}.

\section{Discussion}\label{sec_discussion}
This work presents DT-ICU as a multimodal digital twin for ICU outcome prediction that operates continuously over time and integrates heterogeneous clinical data streams. Beyond achieving strong performance against state-of-the-art baselines, our results show that DT-ICU remains robust under variable input lengths and multitask settings, and that it learns a clinically meaningful structure across modalities. By conditioning on a patient’s accumulated history of observations, DT-ICU constructs an evolving digital representation of the individual patient that supports forecasting and simulation of future clinical trajectories, provides early warning signals for adverse outcomes, and is continuously updated as new observations become available during the ICU stay.

\paragraph{Early and continuous risk estimation.}
Our test-length analysis demonstrates that most discriminative power emerges within the first 12--24 hours after ICU admission, with AUROC and Accuracy rapidly saturating. Notably, even at the earliest evaluation point in our setup (4 hours after admission), the model already achieves meaningful predictive performance, indicating that informative risk signals can be extracted from very limited early data. At the same time, AUPRC continues to improve until around 24--48 hours, suggesting that longer observation windows refine the ranking of truly high-risk patients in a highly imbalanced setting. Clinically, this means that DT-ICU can support early triage while still benefiting from continuous updates as more data become available.

\paragraph{Multimodal structure of ICU reasoning.}
The LOMO and LTMO analyses reveal that different modalities play distinct and complementary roles in prediction. Procedure data, which reflect major clinical interventions, are the dominant drivers of sensitivity. Medications and ingredients provide important complementary signals, especially when paired with procedure information or observations of output signals (urine, drainage, etc.). Charted vitals, dates, diagnoses, and demographic variables act mainly as contextual information but may play a less important role.

This organization closely mirrors how clinicians reason about critically ill patients: interventions signal severity, physiological responses help confirm or rule out risk, and background information provides context but rarely determines decisions in isolation.

\paragraph{Metric behavior and failure modes.}
The strong AUROC--AUPRC correlation across LTMO perturbations implies that multimodal ablations mainly affect the quality of the learned risk ranking, rather than inducing threshold-specific calibration effects.
Precision and Recall exhibit clear trade-offs, revealing two distinct failure modes: removing output-related information leads to overly aggressive predictions with many false positives, whereas removing procedure information produces conservative behavior that misses many true high-risk cases. These complementary roles help explain why jointly removing procedures and outputs causes the most severe overall degradation.

\paragraph{"Learned" multimodal structure.}
The above discussions show that different clinical modalities contribute differently to predictive performance. Intervention information and output observed signals carry the strongest signal, whereas contextual modalities such as dates, charts, diagnoses, and demographics provide complementary but less discriminative information. Importantly, this hierarchy is not imposed by manual feature selection or architectural heuristics. Instead, DT-ICU learns it implicitly through its modality-specific encoders and cross-modal temporal fusion, which allow each data stream to be represented and weighted according to its predictive relevance over time. In this sense, the observed modality importance profiles provide empirical evidence that the digital twin formulation enables the model to discover clinically meaningful structures across heterogeneous inputs, rather than relying on predefined assumptions about their relative importance.

And because this structure is learned rather than predefined, DT-ICU can naturally adapt to heterogeneous patient trajectories and to irregular or missing data streams, instead of relying on fixed rules about which inputs should matter. This allows the model to remain reliable across diverse clinical situations in which data availability and relevance vary over time. Furthermore, this continuous and adaptive formulation makes DT-ICU better aligned with real ICU workflows than one-shot prediction models, supporting early risk estimation followed by continuous reassessment and refinement as new clinical evidence accumulates during the patient’s stay.

\paragraph{Future research directions.}
The present study is evaluated on the MIMIC-IV dataset, which provides a large, publicly available benchmark but represents a single healthcare system; future work should examine the generalizability of DT-ICU across broader clinical settings. While we focus on mortality prediction as a representative task, the digital twin formulation naturally extends to a wide range of clinical endpoints, including disease progression, length of stay, and treatment response, all of which can be framed within classification or regression paradigms. Finally, realizing the full potential of DT-ICU will require prospective evaluation in real-world clinical workflows, in close collaboration with clinicians, to assess its performance on live data and its utility for continuous patient monitoring and case-based decision support.

\section{Methods}\label{sec_methods}

\subsection{Dataset}
\label{sec_dataset}

We conduct experiments on the MIMIC-IV database~\cite{johnson2023mimic}, a large, publicly available collection of de-identified electronic health records from Beth Israel Deaconess Medical Center.
MIMIC-IV contains detailed, time-stamped ICU data, including vital signs, laboratory results, medication administrations, procedures, fluid balance, demographics, diagnoses, etc., making it suitable for multimodal ICU digital twin modeling.
Two versions of MIMIC-IV are used in this work.
For strict comparison with prior literature, we use MIMIC-IV v2.0 when reproducing previously reported baselines that were built on this version. Our main experiments are conducted on MIMIC-IV v3.1, which includes updates and a larger cohort of ICU stays compared with v2.0. Unless otherwise stated, all results in this paper are reported on v3.1.

To ensure consistency with existing work while supporting our multimodal temporal modeling, we use two open-source data extraction pipelines respectively for v2.0 ~\cite{gupta2022extensive} and for v3.1~\cite{mimicICUPreprocess2025}. Both pipelines follow the same extraction logic but are adapted to the schema differences between MIMIC-IV versions. In both cases, raw ICU events are converted into aligned, hourly time series, where all sequential modalities are aggregated into one-hour bins. This produces a unified representation in which medications, procedures, outputs, chart events, date/time features, and ingredient events are synchronized across modalities. Categorical static variables such as gender, insurance type, admission type, ICU type, and race are encoded as integer category labels. These open data extraction pipelines also provide train/val/test splits of the dataset, which we followed in our experiments.
The extracted MIMIC-IV v3.1 consists of $94458$ ICU stays, including $7364$ positive mortality cases (death) and $87094$ negatives (survive), reflecting the severe class imbalance typical of real ICU populations.

Table~\ref{tab:modality_definitions} explains the modalities and events used in our work. When performing strict literature comparisons, we match the feature set of the baselines exactly (for example, excluding admission urgency if not used in that work), and all other reported results use the full set of modalities unless specified.

\begin{table}[ht]
\centering
\small
\caption{Explanation of different events used as inputs.}
\label{tab:modality_definitions}
\begin{tabular}{p{1.6cm} p{7.2cm} p{3.2cm}}
\toprule
\textbf{Abbr.} & \textbf{Description} & \textbf{Source table} \\
\midrule

Meds
& Medications and infusions: administered drugs and fluids with timing and dose
& \texttt{icu.inputevents} \\

Chart
& Bedside charted measurements such as vital signs and monitored variables
& \texttt{icu.chartevents} \\

Out
& Physiological outputs including urine, drains, and other recorded outputs
& \texttt{icu.outputevents} \\

Proc
& Clinical procedures and interventions (e.g., ventilation, invasive support)
& \texttt{icu.procedureevents} \\

Date
& Time-stamped clinical events such as dialysis or procedure timing
& \texttt{icu.datetimeevents} \\

Ing
& Ingredients and intake components of administered inputs (fluids, nutrition, additives)
& \texttt{icu.ingredientevents} \\

Demo
& Demographics and admission information: gender, age, race, insurance, ICU type, admission urgency
& \texttt{patients}, \texttt{admissions}, \texttt{transfers} \\

Diag
& Diagnoses and comorbidities coded during the hospitalization (ICD)
& \texttt{hosp.diagnoses\_icd} \\

\bottomrule
\end{tabular}
\end{table}

\subsection{Evaluation protocol}
We evaluate the performance of the models using standard metrics for imbalanced classification prediction (AUROC, AUPRC, accuracy, precision, and recall),  and mean squared error (MSE) for regression.

Specifically, precision and recall reflect different clinical risks in ICU mortality prediction.
High recall corresponds to correctly identifying as many high-risk patients as possible, whereas high precision reflects a low false-alarm rate.
In this setting, failing to identify a patient who will die (a false negative) is far more dangerous than incorrectly flagging a patient who will survive (a false positive), which mainly leads to additional monitoring or resource use.
As shown in Section~\ref{subsec:lomo_ltmo}, these two quantities cannot be optimized simultaneously and are influenced by different data modalities.
We therefore report both precision and recall throughout and analyze their trade-offs explicitly, rather than collapsing performance into a single scalar metric.
And during training, we explicitly bias the final DT-ICU model toward higher recall while maintaining reasonable precision as missing a truly high-risk patient (false negative) is far more harmful than raising an unnecessary alarm (false positive). As a result, the selected operating point of DT-ICU prioritizes sensitivity to high-risk patients, while precision is used to control alert burden.

\subsection{Problem formulation}
\label{sec_problem}
We model each ICU stay as a partially observed multimodal temporal process.
Let $\mathbf{s}$ denote the static patient attributes, including demographics and diagnoses, and let $\mathbf{x}_t$ denote the collection of all sequential clinical variables observed at time $t$, including medications, procedures, outputs, vital signs, time features, and ingredient events.
Formally, we write
\[
\mathbf{x}_t = \{\mathbf{meds}_t,\; \mathbf{chart}_t,\; \mathbf{out}_t,\; \mathbf{proc}_t,\; \mathbf{date}_t,\; \mathbf{ing}_t\},
\]
where the explanation of each component could be found in Table~\ref{tab:modality_definitions}.

DT-ICU aims to learn the joint conditional predictive model
\begin{equation}
p(\mathbf{x}_{t+1}, r_t \mid \mathbf{x}_{0:t}, \mathbf{s}),
\end{equation}
where $\mathbf{x}_{0:t}$ denotes the full observed multimodal history up to time $t$, $\mathbf{x}_{t+1}$ represents the next-step physiological and treatment-related variables to be predicted, and $r_t$ is the mortality risk at time $t$.
This formulation captures both short-term physiological evolution and instantaneous clinical risk under a unified model, and supports the three essential requirements of a clinically actionable digital twin: First, it enables continuous updating, since all predictions are conditioned on the entire available history $\mathbf{x}_{0:t}$ as new observations arrive.
Second, it supports multimodal and multitask prediction, jointly modeling heterogeneous sequential signals together with static patient attributes to predict both future physiological variables and mortality risk. Third, it allows perturbation-based and forward simulation, because the model can be applied iteratively to explore how changes in input modalities or the absence of certain information affect future predictions.

\subsection{DT-ICU pipeline}
\label{sec_pipeline}
\begin{figure*}[t]
    \centering
    \includegraphics[width=1\linewidth]{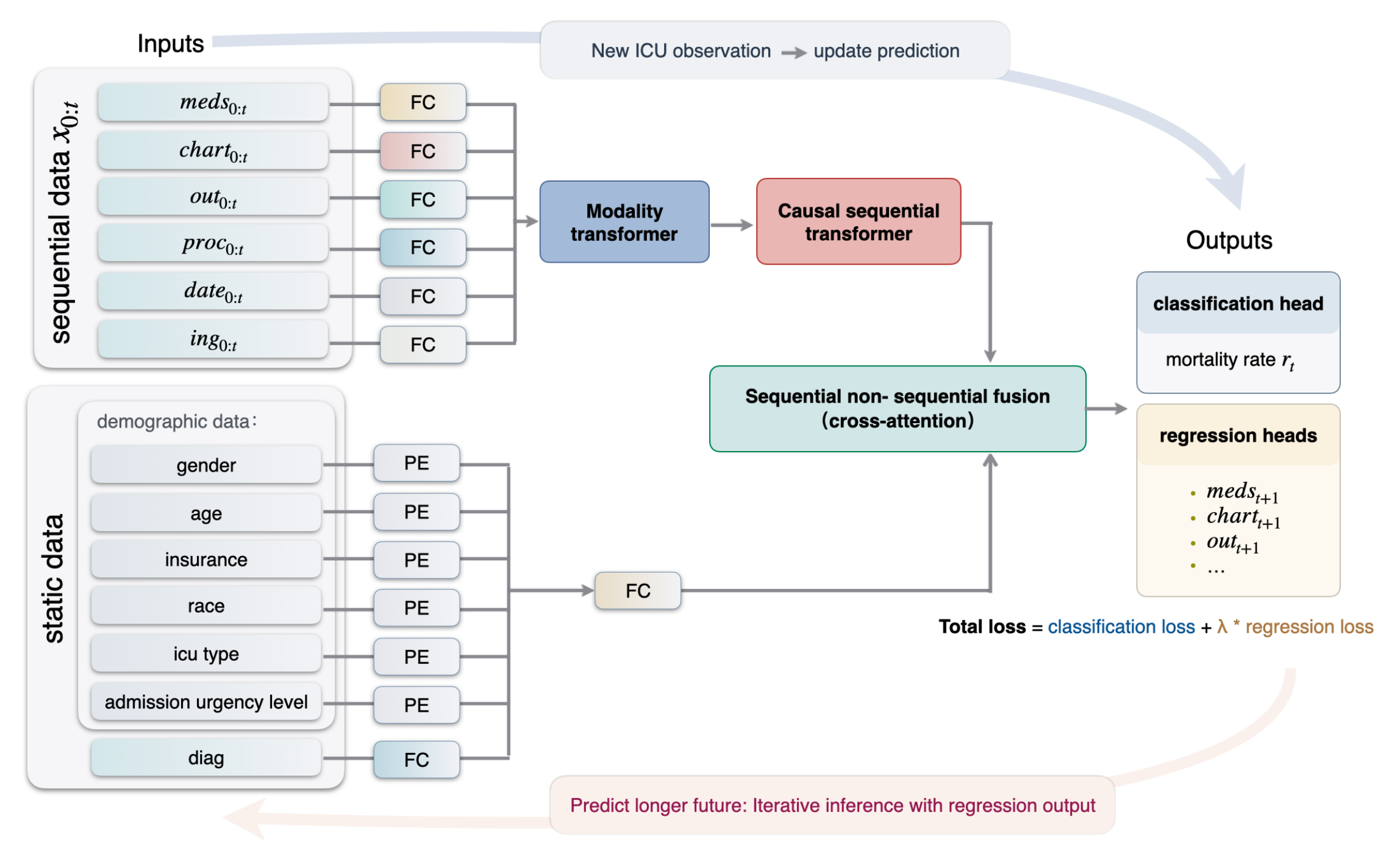}
    \caption{
    Pipeline of our proposed DT-ICU model.
    Multimodal ICU data up to time $t$ are encoded into modality-specific embeddings and processed by a modality transformer followed by a causal temporal transformer.
    Static patient attributes (demographics and diagnoses) are fused with the temporal representation via cross-attention. PE indicates positional encoding, and FC indicates fully connected layers.
    The model outputs both a mortality prediction and next-step physiological forecasts at time $t{+}1$ through multitask heads.
    When new ICU observations become available, predictions could be updated using real measurements.
    When future observations are not yet available, the model rolls out forward in time by feeding its own regression outputs back as inputs, enabling iterative future inference.
    }
    \label{fig:pipeline}
\end{figure*}

Figure~\ref{fig:pipeline} illustrates the full DT-ICU pipeline.
At each time step $t$, the model receives two types of inputs: sequential ICU data $\mathbf{x}_{0:t}$ and static patient information $\mathbf{s}$.
Each sequential modality is first embedded independently through modality-specific fully connected layers.
The resulting embeddings are concatenated and processed by a modality transformer, which learns cross-modal interactions at each time step.
This representation is then passed into a causal temporal transformer, which models longitudinal dependencies using masked self-attention to ensure that predictions at time $t$ depend only on information available up to that time.
Static variables, including demographic features and diagnoses, are embedded separately and projected into a latent representation.
These static embeddings are fused with the sequential representation through a cross-attention module~\cite{vaswani2017attention}, referred to as the \emph{sequential--non-sequential fusion} block, allowing static patient context to modulate temporal dynamics.
The fused representation is then passed to two task-specific heads: a classification head that outputs the mortality risk $r_t$, and a set of regression heads that predict the next-step variables $\mathbf{x}_{t+1}$.
The model is trained with a multitask objective,
\begin{equation}
\mathcal{L} = \mathcal{L}_{\text{cls}} + \lambda \, \mathcal{L}_{\text{reg}},
\end{equation}
where $\mathcal{L}_{\text{cls}}$ is the mortality classification loss, $\mathcal{L}_{\text{reg}}$ is the regression loss for next-step prediction, and $\lambda$ controls their relative weighting.

\paragraph{Continuous updating and iterative inference.}
During deployment, DT-ICU operates in a streaming fashion.
As the model fits any $t$, so when a new ICU observation $\mathbf{x}_{t+1}$ becomes available, it could be directly appended to the historical sequence, forming $\mathbf{x}_{0:t+1}$, and the model immediately recomputes both the mortality risk $r_{t+1}$ and the next-step forecast.
This enables continuous real-time updating as patient data arrives.

When future observations are not yet available, DT-ICU can be used to generate forward simulations via iterative inference.
Starting from the current state $\mathbf{x}_{0:t}$, the model first predicts $\hat{\mathbf{x}}_{t+1}$.
This prediction is then fed back as input, forming an extended sequence $(\mathbf{x}_{0:t}, \hat{\mathbf{x}}_{t+1})$, which is used to predict $\hat{\mathbf{x}}_{t+2}$, and so on.
By recursively applying this procedure, DT-ICU produces a simulated future trajectory together with corresponding mortality risk estimates, enabling counterfactual and what-if analysis of treatment and disease progression.

\subsection{Training setup and evaluation protocol}

All models were trained on a single NVIDIA RTX 4090 GPU. We use the Adam optimizer with a learning rate of $5\times10^{-7}$ and a batch size of 16. Binary cross-entropy loss is used for the mortality classification task, and mean squared error (MSE) is used for the regression tasks. The overall training objective is a weighted sum of these losses,
\[
\mathcal{L} = \mathcal{L}_{\text{cls}} + 0.5\,\mathcal{L}_{\text{reg}},
\]
which encourages accurate mortality prediction while preserving sensitivity to future physiological dynamics.
During training, the input sequence length is randomly sampled between 4 hours and 10 days, reflecting the variable data availability encountered in real ICU settings. This enables the model to learn under heterogeneous temporal contexts and supports continuous operation during inference.

\section{Conclusion}
We presented DT-ICU, a multimodal digital twin framework for continuous ICU outcome prediction that integrates heterogeneous clinical data streams and updates risk estimates over time. Evaluated on MIMIC-IV, DT-ICU achieves strong and stable performance across varying observation lengths and reveals clinically meaningful patterns in how different modalities contribute to prediction. These properties position DT-ICU as a promising foundation for digital twin modeling and monitoring in critical care.

\bmhead{Acknowledgements}
This work was supported by institutional funding and computing resources provided by ETH Zurich.
The author thanks Dr.Diego Paez-Granados, Dr.Oriella Gnarra, Bertram Fuchs and Yanke Li for helpful discussions and assistance regarding dataset selection, data access, and clarifying the importance of explainability in clinical modeling.

\section*{Declarations}
\paragraph{Data availability and ethics.}
All experiments are conducted on de-identified, publicly available data from the MIMIC databases. Ethical approval is not required for secondary analysis of these datasets.

\paragraph{Code availability}
The source code and trained model weights for DT-ICU are publicly available at \url{https://github.com/GUO-W/DT-ICU-release}.

\paragraph{Author contributions}
The author conceived the study, designed the methodology, conducted all experiments, analyzed the results, and wrote the manuscript. Most computational experiments were performed using ETH Zurich's institutional computing resources.

\paragraph{AI-assisted writing disclosure}
Large language models, including ChatGPT, were used to assist with language polishing and grammatical editing of the manuscript. The authors take full responsibility for the content of the manuscript.

\begin{appendices}

\section{Effect of test length}\label{sec_appendix_len}
This appendix reports the complete quantitative results for the test-length study presented in  Figure~\ref{fig:test_length} of Section~\ref{subsec:ablate_test_length}
For each test length, we provide the full set of evaluation metrics, including AUROC, AUPRC, accuracy, precision, recall, and prediction error.
\begin{table}[ht]
\centering
\begin{tabular}{c|ccccc}
\hline
\textbf{Test Length} & \textbf{AUROC} & \textbf{AUPRC} & \textbf{Accuracy} & \textbf{Precision} & \textbf{Recall} \\
\hline
4   & 0.93 & 0.67 & 0.84 & 0.42 & 0.88 \\
8   & 0.95 & 0.73 & 0.90 & 0.56 & 0.87 \\
12  & 0.97 & 0.77 & 0.94 & 0.68 & 0.89 \\
24  & 0.97 & 0.83 & 0.95 & 0.79 & 0.80 \\
36  & 0.97 & 0.82 & 0.95 & 0.76 & 0.84 \\
48  & 0.98 & 0.84 & 0.95 & 0.80 & 0.79 \\
60  & 0.98 & 0.83 & 0.95 & 0.75 & 0.85 \\
72  & 0.97 & 0.82 & 0.95 & 0.76 & 0.85 \\
84  & 0.97 & 0.82 & 0.94 & 0.72 & 0.86 \\
96  & 0.97 & 0.82 & 0.94 & 0.72 & 0.86 \\
\hline
\end{tabular}
\caption{Effect of observation window length on predictive performance}
\label{tab:ablate_test_length}
\end{table}

\section{LOMO and LTMO ablation details}
\label{sec_appendix_lomo_ltmo}

This appendix reports the full numerical results of the LOMO and LTMO ablation experiments used in the multimodal explainability analysis in Section~\ref{subsec:lomo_ltmo}.

For LOMO, Table~\ref{tab:lomo_full} reports both the raw performance of each zeroed-modality configuration and its change relative to the full-modality baseline. We define the change as $\Delta = m_{\text{ablated}} - m_{\text{baseline}}$, such that negative values indicate a degradation in performance when a modality is removed. Reporting both absolute values and relative changes enables direct assessment of each modality’s standalone contribution across all evaluation metrics. The LTMO results are reported in the same unified format, enumerating all leave-two-modalities-out configurations and their corresponding performance shifts. Together, these tables provide a complete and quantitative view of how DT-ICU relies on individual modalities and their interactions, supporting the attribution analyses presented in the main text.
\begin{table}[t]
\centering
\small
\caption{
LOMO ablation results showing both raw performance and change relative to the full-modality baseline.
For each metric we report the ablated performance value followed by the change in parentheses,
where $\Delta = m_{\text{ablated}} - m_{\text{baseline}}$ and negative values indicate performance drops.
}
\label{tab:lomo_full}
\begin{tabular}{lcccccc}
\toprule
\textbf{Zeroed modality} &
\textbf{Loss} &
\textbf{AUROC} &
\textbf{AUPRC} &
\textbf{Accuracy} &
\textbf{Precision} &
\textbf{Recall} \\
\midrule
Baseline &
0.01 (0.00) & 0.98 (0.00) & 0.83 (0.00) & 0.96 (0.00) & 0.62 (0.00) & 0.90 (0.00) \\

Meds &
0.01 (0.00) & 0.98 (0.00) & 0.80 (-0.03) & 0.96 (0.00) & 0.65 (+0.03) & 0.84 (-0.06) \\

Chart &
0.01 (0.00) & 0.98 (0.00) & 0.83 (0.00) & 0.96 (0.00) & 0.62 (0.00) & 0.89 (-0.01) \\

Out &
0.01 (0.00) & 0.98 (0.00) & 0.79 (-0.04) & 0.92 (-0.04) & 0.44 (-0.18) & 0.93 (+0.03) \\

Proc &
0.01 (0.00) & 0.95 (-0.03) & 0.64 (-0.19) & 0.95 (-0.01) & 0.62 (0.00) & 0.60 (-0.30) \\

Date &
0.01 (0.00) & 0.98 (0.00) & 0.83 (0.00) & 0.96 (0.00) & 0.62 (0.00) & 0.90 (0.00) \\

Ing &
0.02 (+0.01) & 0.98 (0.00) & 0.80 (-0.03) & 0.94 (-0.02) & 0.51 (-0.11) & 0.91 (+0.01) \\

Demo &
0.04 (+0.03) & 0.97 (-0.01) & 0.80 (-0.03) & 0.95 (-0.01) & 0.60 (-0.02) & 0.86 (-0.04) \\

Diag &
0.01 (0.00) & 0.98 (0.00) & 0.83 (0.00) & 0.96 (0.00) & 0.62 (0.00) & 0.90 (0.00) \\

\bottomrule
\end{tabular}
\end{table}

\begin{table}[p]
\centering
\scriptsize
\caption{
LTMO ablation results showing both raw performance and change relative to the full-modality baseline.
For each metric we report the ablated value followed by the change in parentheses,
where $\Delta = m_{\text{ablated}} - m_{\text{baseline}}$ and negative values indicate performance drops.
}
\label{tab:ltmo_full}
\begin{tabular}{lcccccc}
\toprule
\textbf{Zeroed pair} &
\textbf{Loss} &
\textbf{AUROC} &
\textbf{AUPRC} &
\textbf{Accuracy} &
\textbf{Precision} &
\textbf{Recall} \\
\midrule
Baseline &
0.01 (0.00) & 0.98 (0.00) & 0.83 (0.00) & 0.96 (0.00) & 0.62 (0.00) & 0.90 (0.00) \\

meds + chart  & 0.12 (+0.11) & 0.98 (0.00) & 0.80 (-0.03) & 0.96 (0.00) & 0.65 (+0.03) & 0.83 (-0.07) \\
meds + out    & 0.21 (+0.20) & 0.97 (-0.01) & 0.76 (-0.07) & 0.92 (-0.04) & 0.47 (-0.15) & 0.91 (+0.01) \\
meds + proc   & 0.19 (+0.18) & 0.91 (-0.07) & 0.49 (-0.34) & 0.94 (-0.02) & 0.64 (+0.02) & 0.32 (-0.58) \\
meds + date   & 0.12 (+0.11) & 0.98 (0.00) & 0.80 (-0.03) & 0.96 (0.00) & 0.65 (+0.03) & 0.84 (-0.06) \\
meds + ing    & 0.15 (+0.14) & 0.97 (-0.01) & 0.79 (-0.04) & 0.95 (-0.01) & 0.60 (-0.02) & 0.82 (-0.08) \\
meds + diag   & 0.12 (+0.11) & 0.98 (0.00) & 0.80 (-0.03) & 0.96 (0.00) & 0.65 (+0.03) & 0.83 (-0.07) \\
meds + demo   & 0.16 (+0.15) & 0.97 (-0.01) & 0.77 (-0.06) & 0.95 (-0.01) & 0.63 (+0.01) & 0.78 (-0.12) \\

chart + out   & 0.23 (+0.22) & 0.98 (0.00) & 0.79 (-0.04) & 0.92 (-0.04) & 0.44 (-0.18) & 0.93 (+0.03) \\
chart + proc  & 0.16 (+0.15) & 0.94 (-0.04) & 0.64 (-0.19) & 0.95 (-0.01) & 0.62 (0.00) & 0.58 (-0.32) \\
chart + date  & 0.13 (+0.12) & 0.98 (0.00) & 0.83 (0.00) & 0.96 (0.00) & 0.62 (0.00) & 0.89 (-0.01) \\
chart + ing   & 0.18 (+0.17) & 0.98 (0.00) & 0.79 (-0.04) & 0.94 (-0.02) & 0.52 (-0.10) & 0.90 (0.00) \\
chart + diag  & 0.13 (+0.12) & 0.98 (0.00) & 0.83 (0.00) & 0.96 (0.00) & 0.62 (0.00) & 0.89 (-0.01) \\
chart + demo  & 0.17 (+0.16) & 0.97 (-0.01) & 0.80 (-0.03) & 0.95 (-0.01) & 0.60 (-0.02) & 0.85 (-0.05) \\

out + proc    & 0.25 (+0.24) & 0.93 (-0.05) & 0.54 (-0.29) & 0.91 (-0.05) & 0.40 (-0.22) & 0.70 (-0.20) \\
out + date    & 0.23 (+0.22) & 0.98 (0.00) & 0.80 (-0.03) & 0.92 (-0.04) & 0.44 (-0.18) & 0.93 (+0.03) \\
out + ing     & 0.34 (+0.33) & 0.97 (-0.01) & 0.77 (-0.06) & 0.87 (-0.09) & 0.34 (-0.28) & 0.95 (+0.05) \\
out + diag    & 0.23 (+0.22) & 0.98 (0.00) & 0.79 (-0.04) & 0.92 (-0.04) & 0.44 (-0.18) & 0.93 (+0.03) \\
out + demo    & 0.34 (+0.33) & 0.97 (-0.01) & 0.76 (-0.07) & 0.89 (-0.07) & 0.36 (-0.26) & 0.93 (+0.03) \\

proc + date   & 0.16 (+0.15) & 0.95 (-0.03) & 0.64 (-0.19) & 0.95 (-0.01) & 0.62 (0.00) & 0.61 (-0.29) \\
proc + ing    & 0.24 (+0.23) & 0.93 (-0.05) & 0.58 (-0.25) & 0.92 (-0.04) & 0.45 (-0.17) & 0.67 (-0.23) \\
proc + diag   & 0.16 (+0.15) & 0.95 (-0.03) & 0.64 (-0.19) & 0.95 (-0.01) & 0.62 (0.00) & 0.60 (-0.30) \\
proc + demo   & 0.21 (+0.20) & 0.93 (-0.05) & 0.60 (-0.23) & 0.95 (-0.01) & 0.64 (+0.02) & 0.52 (-0.38) \\

date + ing    & 0.19 (+0.18) & 0.98 (0.00) & 0.80 (-0.03) & 0.94 (-0.02) & 0.51 (-0.11) & 0.90 (0.00) \\
date + diag   & 0.13 (+0.12) & 0.98 (0.00) & 0.83 (0.00) & 0.96 (0.00) & 0.62 (0.00) & 0.90 (0.00) \\
date + demo   & 0.18 (+0.17) & 0.97 (-0.01) & 0.80 (-0.03) & 0.95 (-0.01) & 0.60 (-0.02) & 0.86 (-0.04) \\

ing + diag    & 0.19 (+0.18) & 0.98 (0.00) & 0.79 (-0.04) & 0.94 (-0.02) & 0.52 (-0.10) & 0.90 (0.00) \\
ing + demo    & 0.27 (+0.26) & 0.97 (-0.01) & 0.77 (-0.06) & 0.92 (-0.04) & 0.44 (-0.18) & 0.89 (-0.01) \\

diag + demo   & 0.17 (+0.16) & 0.97 (-0.01) & 0.80 (-0.03) & 0.95 (-0.01) & 0.60 (-0.02) & 0.84 (-0.06) \\

\bottomrule
\end{tabular}
\end{table}

\end{appendices}

\end{document}